\documentclass[10pt,twocolumn,a4paper]{esaAI}

\usepackage{float}
\usepackage{amsmath}
\usepackage{amsfonts}
\usepackage{amssymb}
\usepackage{courier}
\usepackage{subcaption}
\usepackage{graphicx}
\usepackage[normalem]{ulem}
\usepackage{adjustbox}
\usepackage{textcomp}
\usepackage{soul}
\usepackage{color}
\usepackage{xcolor}
\usepackage{makecell}
\usepackage{url}

\sethlcolor{blue!20}

\title{Fine-tuning LLMs for Autonomous Spacecraft Control: A Case Study Using Kerbal Space Program}

\def\authorEmail{alejandro.carrasco.aragon@alumnos.upm.es}

\author[1]{Alejandro Carrasco\thanks{Corresponding author. E-Mail: \newline \authorEmail}, Victor Rodriguez-Fernandez}
\author[2]{Richard Linares}
\affil[1]{Universidad Politécnica de Madrid, Madrid, Spain}
\affil[2]{Massachussetts Institute of Technology, Massachussetts, USA}

\begin{document}

\makeCustomtitle

\begin{abstract}
Recent trends are emerging in the use of Large Language Models (LLMs) as autonomous agents that take actions based on the content of the user text prompt. This study explores the use of fine-tuned Large Language Models (LLMs) for autonomous spacecraft control, using the Kerbal Space Program Differential Games suite (KSPDG) as a testing environment. Traditional Reinforcement Learning (RL) approaches face limitations in this domain due to insufficient simulation capabilities and data. By leveraging LLMs, specifically fine-tuning models like GPT-3.5 and LLaMA, we demonstrate how these models can effectively control spacecraft using language-based inputs and outputs. Our approach integrates real-time mission telemetry into textual prompts processed by the LLM, which then generate control actions via an agent. The results open a discussion about the potential of LLMs for space operations beyond their nominal use for text-related tasks. Future work aims to expand this methodology to other space control tasks and evaluate the performance of different LLM families. The code is available at this URL: \texttt{https://github.com/ARCLab-MIT/kspdg}.
\end{abstract}

\section{Introduction}
Large Language Models (LLMs) are, without a doubt, the last major breakthrough in the evolution of artificial intelligence systems. Since the release of ChatGPT \cite{chatgpt} at the end of 2022, we have seen a plethora of applications and use cases emerge across various industries. From generating human-like text to aiding in code completion, LLMs have significantly impacted the way we interact with technology and the possibilities of what AI can achieve.

In recent months, the use of LLMs is expanding beyond text-based applications to become \textit{language agents} capable of taking actions based on the context of the system in which they are integrated. By leveraging the contextual information available to them, LLMs can make informed decisions and perform tasks autonomously. 
This new way of creating autonomous agents intersects with the usage of  Reinforcement Learning (RL) algorithms, and provides a way to overcome some of its well-known limitations, such as the sample inefficiency, and the need for a well-defined reward function. Some recent studies have demonstrated how some powerful LLMs, such as GPT-4, can surpass state-of-the-art RL algorithms in complex games just through studying academics texts and reasoning \cite{wu2023spring}, executing sophisticated trajectories and achieving good zero-shot performance.

This work is focused on the domain of space applications and the development of autonomous agents for guidance and control of spacecrafts. In this context, the creation of AI-based agents has mainly been tackled through RL during recent years, and in fact, we can find RL-based agents for different tasks such as sensor-tasking \cite{siew2022space} and planetary landing \cite{gaudet2020deep}. However, unlike other AI research areas, the space domain lacks of publicly available simulation environments, which are crucial for training AI agents in complex space operations and providing a standard benchmark for evaluating different AI and autonomous control methods. To address this issue, Allen et al. introduced \textit{SpaceGym} \cite{10115968}, a set non-cooperative game environments that are intended to spur development and act as proving grounds for autonomous and AI decision-makers in the orbital domain. Among the available environments in \textit{SpaceGym}, in this work we focus on the Kerbal Space Program Differential Games suite (KSPDG). KSPDG is a suite of differential games, such as pursuit-evasion scenarios, encoded within the Kerbal Space Program (KSP) game engine \footnote{https://www.privatedivision.com/portfolio/kerbal-space-program/} and standardized with OpenAI Gym \cite{brockman2016openai} and PettingZoo \cite{terry2021pettingzoo} interfaces, facilitating the use of diverse AI techniques, including multi-agent reinforcement learning.

While KSPDG presents an innovative framework for testing AI and autonomous control methods in space applications, it is unsuitable for RL training, due to technical and non-technical reasons. On the one hand, the KSP engine, which underpins KSPDG, lacks the capacity for the parallel, accelerated, and headless operations essential for extensive faster-than-real-time RL training. On the other hand, the principled stance of KSPDG's creators to focus on evaluation rather than training emphasizes the need for a ``true test set" environment where overfitting is minimized, and the genuine and unbiased capabilities of AI agents are tested. This approach diverges from the typical RL methodology that relies on iterative training and fine-tuning of agents within a specific simulation environment.

\begin{figure}[hbt]
    \centering
    \includegraphics[width=.95\columnwidth]{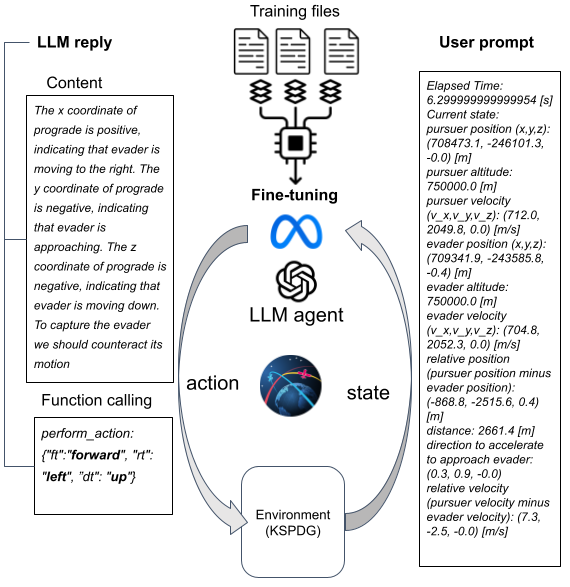}
  \caption{Overview of the proposed approach to use a fine-tuned LLM (e.g. ChatGPT, LLaMA) as an autonomous spacecraft operator that gets, as user prompt, the current status of the mission from the KSDPG simulation environment (i.e., the state or observation in the RL jargon), and replies with a reasoned action to carry out, expressed as a function calling with the specific throttle vector and the textual justification behind the action.}
  \label{fig:overview}
\end{figure}

To overcome the limitations of RL in creating autonomous agents for environments such as KSDPG, as well as for other space operations where numerous simulated data cannot be provided, we propose to adapt the current trend of LLM-based agents to develop an ``intelligent" operator that controls a spacecraft based on the real-time telemetry of the environment, using language exclusively as the input and output of the system. As depicted in \cref{fig:overview}, we design the classic RL loop by interfacing the simulation environment (KSDPG) with a LLM, transforming the real-time observations (or state) of the mission as textual user prompts that are fed to the model. The LLM then processes the prompt and replies with an action that will be plugged in KSDPG to control the spacecraft. Our agent was ranked 2nd in the KSPDG challenge \footnote{https://www.ll.mit.edu/conferences-events/2024/01/kerbal-space-program-differential-game-challenge}, and was presented via a live demonstration during a special session at AIAA SciTech in January 2024.

In our previous study \cite{rodriguez-fernandez2024language}, prompt engineering was the primary focus, and LLM models demonstrated outstanding performance with zero-shot and few-shot prompts. That work utilized prompt engineering to effectively control a spacecraft in the Kerbal Space Program (KSP) simulation, achieving exceptional but non-generalizable results. Additionally, some fine-tuning experiments were conducted using the OpenAI fine-tuning API. At the time, the latest and most powerful model available for fine-tuning was \textit{gpt-3.5-turbo-0125}. The customization of this fine-tuning process relied mainly on the data and three hyperparameters: number of epochs, batch size, and learning rate multiplier \cite{openai_finetuning}. These experiments achieved moderate results due to the API's limitations and its economical cost of fine-tuning.

The objective of the current study is to expand upon previous research by incorporating a set of open-source tools to overcome the limitations of the earlier framework. In contrast to the previous paper, this study focuses solely on the PE1\_I3\_E3 scenario. PE refers to the pursuer-evader problem (rendezvous), I3 denotes the initial position (2.7 km of separation distance), and  E3 represents a heuristic maneuvering technique. \cite{kspdg}. In the past two years, the landscape of LLMs has undergone significant transformations, characterized by frequent and substantial updates. Given these advancements, the "mighty" ChatGPT model is now rivaled by capable models such as Claude\cite{claudeAI}, Gemini\cite{geminiAI}, and open-source models like Mistral\cite{mistral2023mistral7b} and LLaMA\cite{touvron2023llama}. This research focuses on these latter models due to their enhanced flexibility and open source nature, which is crucial for pioneering research in this relatively unexplored area.


The integration of a code agent with KSP is facilitated through a Remote Procedure Call (RPC) program that connects to the selected environment within the game. After each state update, the agent is able to execute an action from a defined set of continuous throttle commands. For a more verbose interaction with the LLM, the actions are verbal—forward, backward, right, left, up, and down—which are then converted into full throttle, full reverse throttle, or no action for each of the three thrusts. This discretization also allows the model to be more like a `human pilot' instead of a control algorithm.

\begin{figure}[t]
    \centering
    \includegraphics[width=.95\columnwidth]{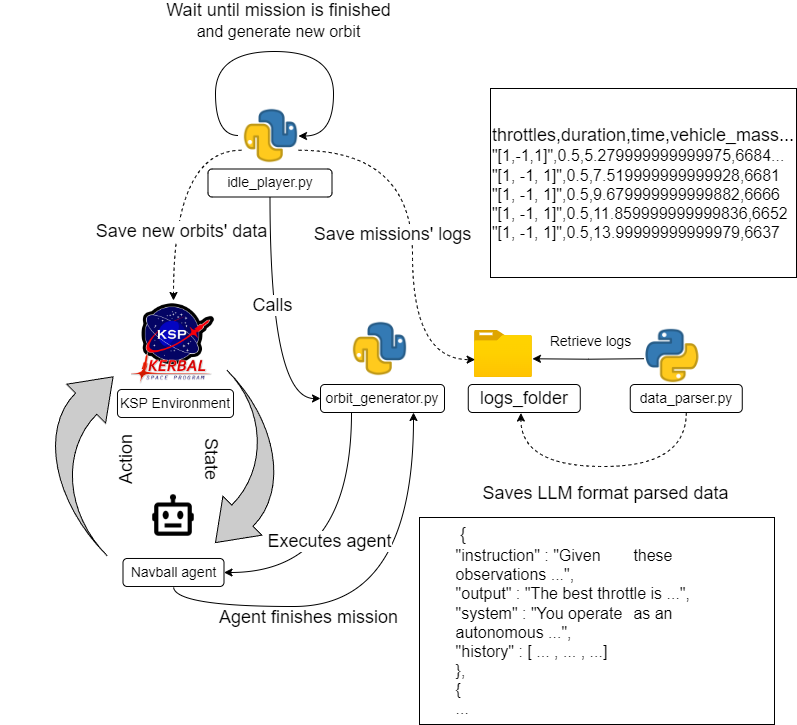}
  \caption{Diagram of the data generation process for fine-tuning the model. The sequence is as follows: (1) The orbit generator is invoked to create a new orbit. (2) The new orbit is saved into KSP. (3) The navball agent is activated to navigate the orbit and generate logs. (4) The logs are saved by the orbit generator. (5) After sufficient runs (e.g., 100), the script data parser converts the logs into text suitable for LLM processing.}
  \label{fig:orbit_generation}
\end{figure}

The primary challenge to address when fine tuning a model for KSPDG is the unavailability of varied mission scenarios, and the lack of expert gameplay logs completing those missions. In fact, this is one of the keys that prevented us from scaling up fine-tuning in \cite{rodriguez-fernandez2024language}. \cref{fig:orbit_generation} depicts a diagram of a program, alongside a bot that tracks the KSP navball's information\footnote{https://wiki.kerbalspaceprogram.com/wiki/Navball} and aligns the vessel to its prograde, to generate a number of pairs of randomized orbits for the pursuer and the evader problem. 

The eccentricity, inclination, semimajor axis, and true anomaly of the pursuer’s orbit were randomly generated within the given constraints: eccentricity $\leq$ 0.1, inclination within 5 degrees of the evader’s orbit, and an initial distance $\leq$ 3 km. The longitude of the ascending node and argument of periapsis were kept constant, ensuring the mission feasibility within a 4-minute span.


Once the issue of data availability is resolved, the real focus of this research—fine-tuning—can be tackled. While OpenAI models, such as those used in previous studies, offer high-quality function calling and perform well with minimal customization, they present significant limitations for large-scale training due to cost and reduced flexibility. These constraints limited our previous experiments to using only 1-2 files of human gameplay logs with discretized throttle actions to test the behavior of GPT models with limited data. Though these initial results were promising, they underscored the need for a more adaptable and cost-effective solution.


To address the challenges in this study, which demand specialized domain-specific knowledge for mission success, we transitioned to an open-source model capable of local training. LLaMA, the most renowned open-source model as of 2024, was chosen for its broad adoption and extensive research community support. The flexibility and adaptability of LLaMA allow for comprehensive fine-tuning tailored to our specific needs, moving beyond aggressive prompt engineering to a more data-driven approach.

To fine-tune LLaMA efficiently\footnote{The LLaMA version used in this work is LLaMA-3-8B}, we utilized a single workstation equipped with five RTX 4090 GPUs, employing several optimization techniques and tools to enhance efficiency:

\begin{itemize}
\item \textbf{Low-Rank Adaptation (LoRA)}: LoRA reduces the number of trainable parameters by factorizing weight updates into low-rank matrices. Being more computational effective \cite{hu2024lora}.
\item \textbf{Hugging Face Transformers Library}: This library facilitated the management of model architecture and training processes \cite{wolf2020transformers}.
\item \textbf{Quantization}: To reduce model size and enhance inference speed, we applied quantization, converting weights and activations to lower precision without significantly impacting performance \cite{nagel2021white}.
\item \textbf{LLaMA Factory}: This tool streamlined the fine-tuning process by efficiently managing datasets, pre-processing, and training tasks \cite{zheng2024llamafactory}.
\end{itemize}

Training hyperparameters included a batch size of \texttt{2}, a learning rate of \texttt{1e-4}, a cosine learning rate scheduler, \texttt{3} epochs, a LoRA configuration (\texttt{rank=16, alpha=8, dropout=0.05}), Flash Attention 2 and Dora enabled, and gradient accumulation steps of \texttt{2}. Fine-tuning only the last layers (\texttt{2} in this case) and using a small batch size are recommended \cite{Masters2018RevisitingSB, Liu2021AutoFreezeAF}. The very small batch size here was due to hardware limitations. Extending the training beyond \texttt{3} epochs did not improve loss.

The LLaMA dataset consisted of 50 top-performing randomly generated orbit missions from the navball agent (see \cref{fig:orbit_generation}), chosen for their optimal distance (meters) and approach speed (seconds). These missions employed less aggressive prompting while still utilizing the chain-of-thought technique \cite{wei2022chain} to enable the model to learn and apply its own reasoning effectively.

\section{Results}

In \cref{tab:fine_tune_2}, we present the results of fine-tuned GPT models using human gameplays and LLaMA models using navball agent gameplays. GPT models show gradual improvement, surpassing the baseline after two training gameplays, indicating the potential of a fine-tuned model with more gameplays. Note that the simple fine-tuning in the GPT experiments used only one file, basic prompting, and the default hyperparameters selected by the OpenAI API. The LLaMA dataset was divided into subsets of 10, 25, and 50 gameplay files. One subset—the 10 files—utilized a sliding window technique, where previous actions were included to provide the model with additional context. LLaMA models consistently exceed their baseline performance, where the closest distance is \textasciitilde 120 meters better than GPT's baseline\footnote{The GPT Model is 3.5, which is older than LLaMA 3}. The best LLaMA models perform exceptionally well, demonstrating the potential benefits of larger datasets, indicating the potential of a fine-tuned model with more gameplays. Finally, the model utilizing the sliding window technique demonstrates great performance results leveraging the context of the LLM.

\begin{table}[htb]
\renewcommand{\arraystretch}{1}
\begin{center}
\begin{adjustbox}{width=0.5\textwidth}
\begin{tabular}{l || p{1.2cm} | p{1.2cm} | p{1.2cm} | p{1.5cm} | p{2.2cm} }
\hline\hline
\textbf{Method} & \multicolumn{3}{|c|}{\makecell{\textbf{Distance (m)} \\ \textbf{Best} \hspace{0.5cm} \textbf{Average} \hspace{0.5cm} \textbf{Worst}}} & \makecell{\textbf{Failure} \\ \textbf{Rate}} & \makecell{\textbf{Average} \\ \textbf{Latency (ms)}} \\
\hline\hline
baseline GPT & 178.11 & 200.10 & 232.16 & 36.8\% & 840.42 \\
simple fine-tuning & 263.55 & 265.89 & 271.51 & \textbf{0.0\%} & 987.43 \\
+ hyperparameter tuning & 188.90 & 202.08 & 210.62 & \underline{0.1\%} & \dashuline{831.30} \\
+ system prompt & 197.41 & 214.87 & 227.67 & \textbf{0.0\%} & \underline{753.52} \\
+ two train gameplays & 132.09 & 159.78 & 200.47 & \dashuline{0.2\%} & \textbf{557.49} \\
\hline
baseline LLaMA & 52.69 & 140.68 & 267.32 & 9.09\% & 8580.43 \\
fine-tune 10 files & 30.52 & 51.53 & 80.88 & \textbf{0.00\%} & 3444.88 \\
fine-tune 25 files & \dashuline{13.54} & \underline{29.44} & 58.83 & \textbf{0.00\%} & 3316.89 \\
fine-tune 50 files & \underline{11.86} & \dashuline{29.76} & \dashuline{48.81} & \textbf{0.00\%} & 3455.29 \\
fine-tune 10 files win=3 & 23.08 & 40.03 & 49.28 & \textbf{0.00\%} & 3292.44 \\
\hline
human gameplays     & \textbf{5.97} & \textbf{6.25} & \textbf{6.54}& - & - \\
navball agent         & 34.34 & 36.43 & \underline{39.76} & - & - \\
\hline\hline
\end{tabular}
\end{adjustbox}
\caption{Performance of fine-tuned models for each technique, measured in distance (meters) and latency (milliseconds). \textbf{Bold} indicates best, \underline{underline} indicates second best, and \dashuline{dashed underline} indicates third best.}
\label{tab:fine_tune_2}
\end{center}
\end{table}

\begin{figure}[t]
    \centering
    \includegraphics[width=.95\columnwidth]{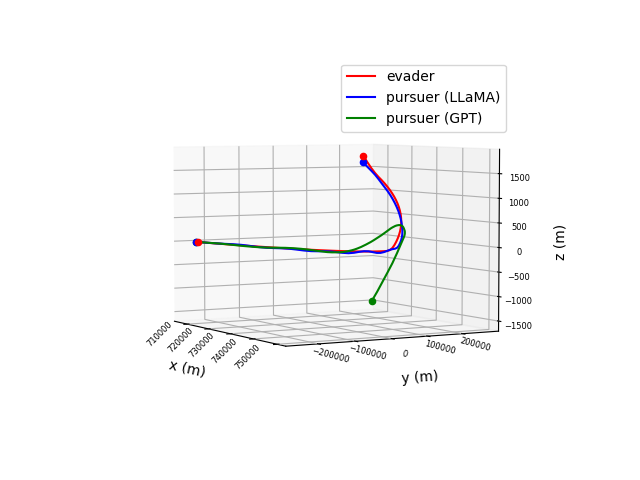}
  \caption{This 3D plot depicts the trajectories of the best-performing fine-tuned models for GPT and LLaMA, along with the evader's path. Due to an incorrect hint, the GPT model deviates significantly after overshooting its target, while the LLaMA model maintains a closer trajectory to the evader.}
  \label{fig:fine-tuning-trajectories-results}
\end{figure}

The fine-tuning trajectories in \cref{fig:fine-tuning-trajectories-results} indicate that the data ingested by the model aids in understanding the problem and determining appropriate actions. However, these trajectories also show that the model's prior knowledge and reasoning still influence its performance. For instance, an incorrect hint, as depicted by the GPT trajectory, deteriorates the model's performance and makes the agent recede from the evader once it overshoots (meaning when it goes past the evader). In contrast, an "agnostic" prompt that complements rather than dictates the model's reasoning can even surpass the dataset results.

\section{Discussion}
The preliminary results of this study demonstrate that LLMs possess the capability to perform technical tasks beyond merely generating verbose text. Fine-tuning these models enhances their reasoning for autonomous space control missions without solely depending on the hints and reasoning provided in the prompt. This process yields a generalized model that can interact as an agent in KSP rendezvous missions.

The fine-tuned LLaMA model clearly surpasses the navball agent results, even in average distance, which poses a singular use case where the model outperforms the agent responsible of creating its own training data. The tests were run on validation sets, thus remarking the generalization capacity LLMs offer.

However, the complexity of a trajectory is not solely determined by distance. This highlights the necessity for more complex and dynamic scenarios that require additional metrics. This issue will be addressed in the next stages of this research.

Moreover, the close distances achieved by the LLaMA models pave the way for exploring docking operations, where the increased complexity will test the robustness of the model's chosen trajectories.

We also plan on leveraging the advantages offered by multi-modal LLMs, such as the recently released GPT-4o \cite{openai_gpt4o} (where 'o' stands for 'omni') and the Phi-3 family \cite{microsoft_phi3}, an open-source model incorporating multi-modal capabilities. Specifically, we intend to utilize vision capabilities in conjunction with language, as both modalities show the potential for creating an agent with human-like decisions.

\bibliographystyle{unsrt}  
\bibliography{library}  

\end{document}